\documentclass{article}

\usepackage{arxiv}
\usepackage{amssymb}
\usepackage{algorithm}
\usepackage{algpseudocode}
\usepackage[utf8]{inputenc} 
\usepackage[T1]{fontenc}    
\usepackage{url}            
\usepackage{booktabs,footnote}       
\usepackage{amsmath,amsfonts}       
\usepackage{nicefrac}       
\usepackage{microtype}      
\usepackage{lipsum}
\usepackage{graphicx}

\algdef{SE}[SUBALG]{Indent}{EndIndent}{}{\algorithmicend\ }%
\algtext*{Indent}
\algtext*{EndIndent}

\title{A Reinforcement Learning Approach for Process Parameter Optimization in Additive Manufacturing}

\author{
 Susheel Dharmadhikari \\
  Department of Mechanical Engineering\\
  Pennsylvania State University\\
  State College, PA 16802 \\
  \texttt{sud85@psu.edu} \\
   \And
 Nandana Menon \\
  Department of Mechanical Engineering\\
  Pennsylvania State University\\
  State College, PA 16802 \\
  \texttt{nfm5316@psu.edu} \\
  \And
 Amrita Basak \\
  Department of Mechanical Engineering\\
  Pennsylvania State University\\
  State College, PA 16802 \\
  \texttt{aub1526@psu.edu} \\
}

\begin{document}
\maketitle



\begin{abstract}
Process optimization for metal additive manufacturing (AM) is crucial to ensure repeatability, control microstructure, and minimize defects. Despite efforts to address this via the traditional design of experiments and statistical process mapping, there is limited insight on an on-the-fly optimization framework that can be integrated into a metal AM system. Additionally, most of these methods, being data-intensive, cannot be supported by a metal AM alloy or system due to budget restrictions. To tackle this issue, the article introduces a Reinforcement Learning (RL) methodology transformed into an optimization problem in the realm of metal AM.  An off-policy RL framework based on Q-learning is proposed to find optimal laser power ($P$)- scan velocity ($v$) combinations with the objective of maintaining steady-state melt pool depth. For this, an experimentally validated Eagar-Tsai formulation is used to emulate the Laser-Directed Energy Deposition environment, where the laser operates as the agent across the $P-v$ space such that it maximizes rewards for a melt pool depth closer to the optimum. The culmination of the training process yields a Q-table where the state ($P,v$) with the highest Q-value corresponds to the optimized process parameter. The resultant melt pool depths and the mapping of Q-values to the $P-v$ space show congruence with experimental observations. The framework, therefore, provides a model-free approach to learning without any prior. 

\end{abstract}



\section{Introduction}

A major challenge to the wider acceptance of metal additive manufacturing (AM) is the lack of a streamlined process qualification methodology. While metal AM boasts of an array of advantages such as reduced material, low energy, low cost of prototyping, and reduced number of parts, metal AM processes suffer from the difficulty of maintaining repeatability~\cite{debroy2018additive}. This is essential for mission-critical components since metal AM is chiefly employed in the manufacture and repair of high-impact, high-cost components and materials in aerospace and gas turbine industries~\cite{ferster2018effects}. Process inconsistencies in metal AM can be attributed to the complex physics involved in the processing of different alloys and AM systems. This necessitates a proper knowledge-backed consensus on these methodologies to avoid dimensional inaccuracies and defects.

Process parameter optimization is therefore an important procedure that needs to be performed to characterize any alloy and/or metal AM system to determine the operable window yielding the desired deposit characteristics. While multiple parameters are at play in a metal AM system, laser power ($P$) and scan velocity ($v$) are easily modified within the system's process window. The individual and combined effect of these parameters can control the geometric, microstructural, and mechanical properties. For example, linear energy density, defined as $\frac{P}{v}$, controls the cooling rates and hence the grain size; a higher energy density garners finer microstructures due to slower cooling rates~\cite{farshidianfar2016effect}. Quite often, the melt pool generated during the metal AM process is employed as a proxy for the thermal and microstructural signatures~\cite{vasinonta2000process,basak2016additive}. Inconsistent melt pool depths across layers and tracks can result in geometric inconsistencies, keyholing, and lack of fusion defects. Therefore, to maintain microstructural and mechanical integrity, seeking optimal process parameters is necessary.

Accordingly, there has been a consistent effort towards finding optimal set of parameters. Traditionally, online monitoring has been employed for process control where process parameters have been modified using feedback control strategies~\cite{berumen2010quality,raghavan2013heat, everton2016review, liao2022simulation}. Researchers have explored parametric studies to gain an understanding of the process-parameter-property relationship to make an informed decision on the optimal set before starting the processing/ optimization. Gockel et al.~\cite{gockel2013understanding} explored the ability to predict and control as-deposited microstructure via process maps for electron beam additive manufacturing of Ti-6Al-4. These process maps were used to identify favourable regions. A comparison of the microstructure process map to the corresponding map for  melt pool dimensions was proposed to determine the relationships that can be used to indirectly control solidification. Bhardwaj et al.~\cite{bhardwaj2019direct} used response surface methodology experimental design technique and ANOVA for the process optimization of track dilution in the L-DED of Titanium-Molybdenum alloy.  Basak et al. \cite{basak2016additive} explored a design of experiments-based approach for optimizing scanning laser epitaxy parameters for CMSX-4\textsuperscript{\textregistered}. Aboutaleb et al.~\cite{aboutaleb2017accelerated} used experimental data from literature to generate the next batch of optimal experiment parameters by modeling the difference in responses between the prior and current experimental data. However, such methods and the corresponding data are subjective to the concerned experiments and may fail when the material and/or metal AM systems involved operate at different processing windows. 

A more favorable approach towards optimization has been through the widely explored supervised machine learning and statistical techniques. Velázquez et al.~\cite{velazquez2021prediction} used analysis of variance and fuzzy interference to identify the impact of process parameters on various bead geometry characteristics. Lu et al.~\cite{lu2010prediction} employed least squares support vector machines and neural nets to find a process-parameter-property-relationships for laser directed energy deposition of SS316L. The recently developed physics-aware MeltPoolNet~\cite{akbari2022meltpoolnet} on the extensive experimental data from over 80 articles claims to serve as a benchmark for melt pool control and optimization. It is a combination of several machine learning (ML) techniques such as  random forests, Gaussian processes, support vector machines, linear regression, gradient boosting, neural nets, and logistic regression. However, a common concern towards the development of all such techniques is the need for large amounts of initial data for reliable predictions. For AM systems that often involve expensive data generation, the use of such techniques is restrictive. 

A better alternative to these data-intensive tools is through algorithms that rely on \textit{active} or \textit{experience} learning, i.e. developing a strategy during the data generation process. Such techniques, while learning on-the-fly, direct the data generation process through better sampling thereby eliminating unnecessary exploration. Furthermore, due to continuous learning, depending on the budget constraints, a feasible algorithm can be constructed at any stage in the process with varying levels of fidelity. Bayesian optimization is one such example of \textit{active learning} algorithms that have been scarcely applied for process optimization with minimal data. Mondal et al.~\cite{mondal2020investigation} developed Gaussian Process surrogate-based Bayesian Optimization for temporal control of melt pool depth using data from an experimentally validated analytical function.
Menon et al.~\cite{menon2022multi} extended the work by adding a finite element perspective to the analytical function and implemented a multi-fidelity Gaussian Process based surrogate to perform Bayesian Optimization. Wang et al.~\cite{wang2019data} developed a surrogate-based parameter optimization with uncertainty quantification using physics-informed computer simulations for electron beam melting of Ti-6Al-4V. Ye et al.~\cite{ye2018novel} combined stochastic finite element analysis with particle swarm optimization for directed energy deposition.

A contemporary competitor to these algorithms has emerged in the form of reinforcement learning (RL) \cite{kaelbling1996reinforcement}. As opposed to \textit{active learning}, RL employs a form of \textit{experience learning} that involves policy formulations while interacting with the concerned environment. It has steeply gained prominence due to successful solutions to problems that had collectively suffered with the aforementioned traditional algorithms, particularly with robotics \cite{kober2013reinforcement} and gaming applications \cite{kaiser2019model, silver2018general}. Most recently, RL has been the center of immense discussion due to its success in improving on a 50-year-old matrix multiplication solution using Alpha Tensor \cite{fawzi2022discovering}. To the AM community, RL is particularly appealing owing to one of its model-free approaches to learning which eliminates the need for any information collection prior to its usage. The percolation of the novel techniques in RL to the relatively nascent field of AM has been understandably limited. Mostly, in AM, RL has been applied for toolpath optimization~\cite{mozaffar2020toolpath}. There have been fewer applications of RL to control of dynamic process parameters to mitigate defects \cite{ogoke2021thermal, zimmerling2022optimisation}. However, the current literature pool does not have any studies on the use of RL for steady-state process parameter optimization that can aid in either calibrating new AM equipment or novel experiments for a desired melt pool depth. The perusal of literature, therefore, indicates that a combination of RL with process parameter optimization is a rich and untested domain that can inevitably benefit the AM community.

To that end, this article proposes the model-free, off-policy RL algorithm - Q-learning~\cite{watkins1992q} for process parameter optimization for melt pool depth within the $P-v$ domain of a candidate AM system i.e., powder-fed laser directed energy deposition (L-DED). The article shows a way to morph the process optimization problem into the Q-learning architecture by uniquely representing the interactions between an \textit{agent}, \textit{action space}, \textit{state space}, and the \textit{environment}. As a proof of concept, an experimentally validated Eagar-Tsai formulation is used as the environment. The optimal $P-v$ combinations obtained from the Q-learning algorithm, when mapped to the experimental observations, reveal a deviation within $0.05 \ mm$ for a desired melt pool depth of $1 \ mm$. Further exploration of the algorithm is conducted through a study of five hyperparameters, viz. (i) domain discretization, (ii) exploration-exploitation tradeoff parameter, (iii) discount factor, (iv) learning rate, and (v) number of episodes, to study their individual influence. This is followed by a discussion of the limitations of the algorithm with some recommendations to tackle the shortcomings in the future.

\begin{figure}[!h]
    \centering
    \includegraphics[width=0.9\textwidth]{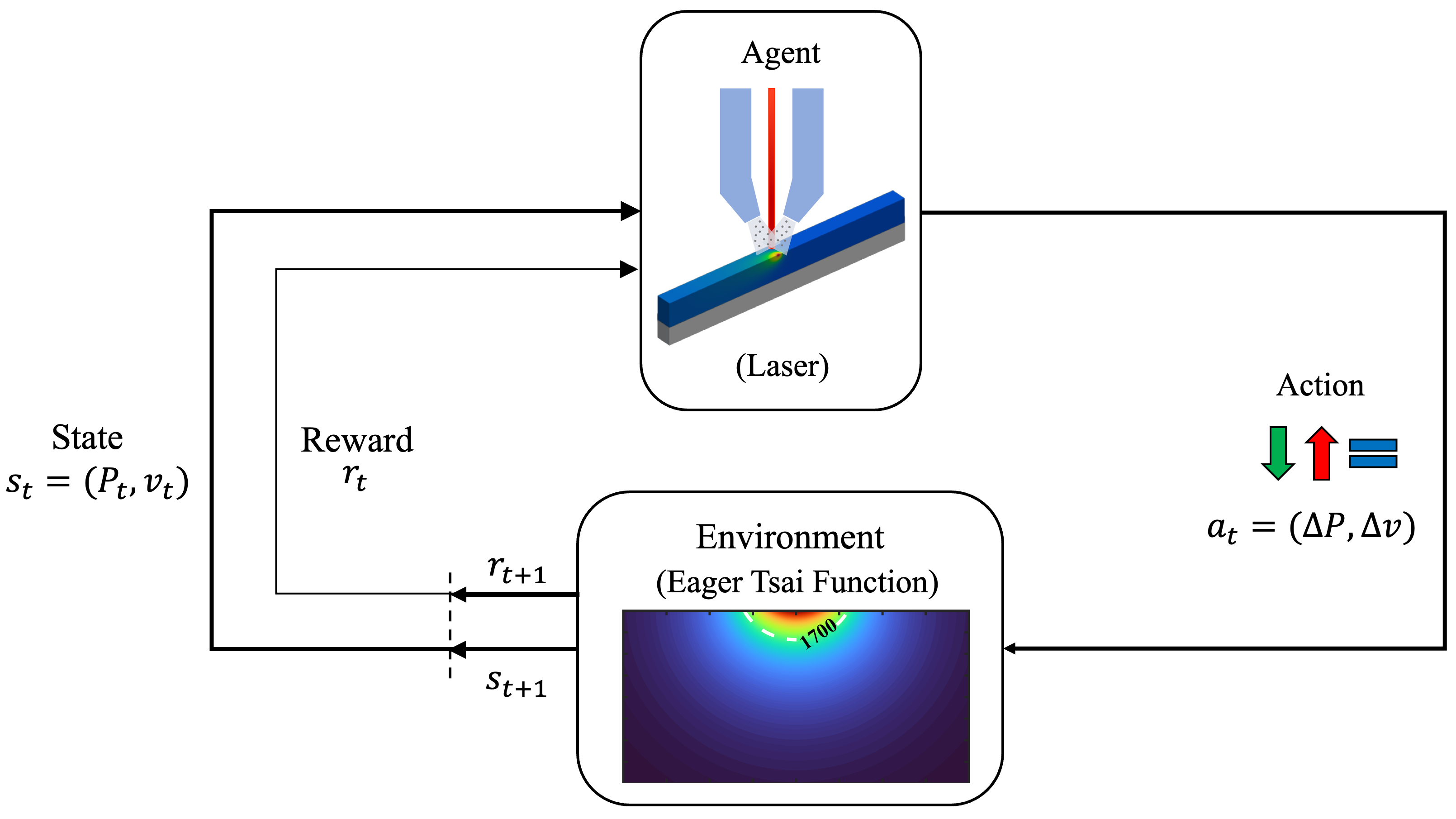}
    \caption{The agent-environment interaction of the traditional RL framework adapted to optimize process parameters in L-DED. The environment is emulated using the Eagar-Tsai function. The laser acts as the agent. The states ($s_t$) correspond to discrete $P_t, v_t$ from the predefined $P-v$ space. The actions ($a_t$) correspond to a change in $P$ and $v$.}
    \label{fig:rl_fw}
\end{figure}

\section{Methodology}
\subsection{Process parameter optimization and Q-learning}
Typically, to fabricate defect-free AM components, the user (or manufacturer) needs to calibrate the metal AM system for an array of materials. This involves selecting combinations of process parameters ($P, v$) from a pre-established process parameter space such that a desired property (e.g., a steady-state melt pool depth of $1\ mm$) is obtained. Moreover, there is no prior information available about the system. This is essentially an optimization problem albeit without the conventional system model that aids in computing the solution. The traditional and inevitable approach, under such circumstances, is often by employing design-of-experiments based sampling and investigation through the parameter space. However, due to the expensive nature of metal AM systems and alloys, such trial-and-error based techniques are prohibitive and extraneous. Evidently, a systematic approach to efficiently \textit{sample} and \textit{learn} needs to be established. To that end, the architecture of the Q-learning framework provides precisely the setup to tackle such problems. Q-learning is one of the simpler methods from the myriad tools in the reinforcement learning literature~\cite{watkins1992q}. It is particularly attractive for this problem due to its \textit{model-free} approach that can provide a pathway to optimization with no prior information. Such an algorithm can thus be used to guide process optimization without an initial understanding of the $P-v$ space for any material or metal AM system. The Q-learning algorithm can be visualized as an interaction between an agent and an environment, as shown in Figure. \ref{fig:rl_fw}. 

\begin{figure}[!h]
    \centering
\includegraphics[width=0.75\textwidth]{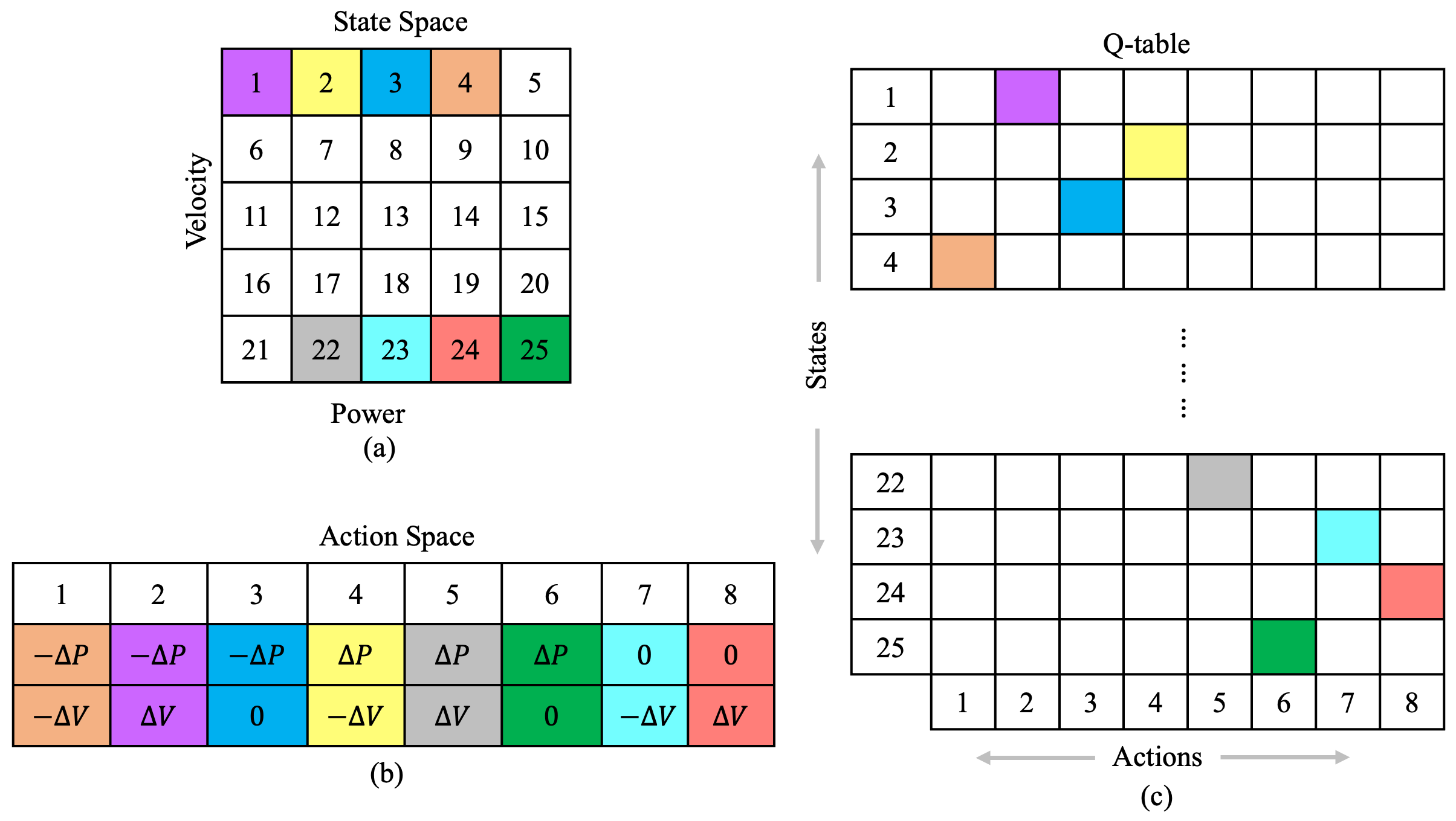}
    \caption{An illustration of the (a) state space, (b) action space, and (c) Q-table. Here, the state space is discretized into a $5 \times 5$ $P-v$ domain. The action space consists of eight unique actions. This yields a $25 \times 8$ dimensional Q-table. The color combinations demonstrate the mapping of the state and action space to the Q-table. For example, a combination of the second state (yellow) and fourth action maps to the corresponding (yellow) location on the Q-table.}  
    \label{fig:qrep}
\end{figure}

In the current context, the environment is assumed to be an L-DED system (simulated via the Eagar-Tsai formulation), and the agent is the laser. The agent can operate within the state space $\mathbb{S}$ that consists of the $P-v$ discretized domain (Figure. \ref{fig:qrep}(a)). Consider that, at time $t$, the agent is in state $s_t$. To move to a new state, the agent can take an action ($a_t$) from the action space $\mathbb{A}$. For the current problem, an action consists of a simultaneous change in $P$ and $v$ that can have eight different combinations as shown in Figure. \ref{fig:qrep}(b). The action taken by the agent is either rewarded or penalized ($r_t$) based on the difference of the $\delta$ evaluated at the state, from the desired depth ($\delta_{\text{opt}}$). Thereafter, the movement through the state space is guided in a way that the agent is able to maximize the total rewards. The quality of each state-action pair is measured in terms of the \textit{Q-value} recorded in a Q-table. This Q-table is a composite representation of the state-action space as shown in Figure. \ref{fig:qrep}(c). The Q-values are updated using the following equation~\cite{sutton2018reinforcement}:

\begin{equation} \label{eq}
    {\underbrace{Q^{new}(s_t, a_t)}_{\text{\normalfont{Updated Q-value}}} \leftarrow \underbrace{Q(s_t, a_t)}_{\text{\normalfont{Current Q-value}}} + \underbrace{\alpha}_{\text{\normalfont{Learning rate}}} \quad \left[ \underbrace{r_t}_{\text{\normalfont{Reward}}} + \underbrace{\gamma}_{\text{\normalfont{Discount factor}}} \underbrace{(\max_{a} Q(s_{t+1}, a))}_{\text{\normalfont{Future optimal Q-value}}} \quad - \quad Q(s_t, a_t) \quad \right]}
\end{equation}

Based on this equation, at the time step \textit{t}, the updated Q-value is a combination of the current value and a future value weighted by a learning factor $\alpha$, and a discount factor $\gamma$. A detailed workflow is explained via the Algorithm \ref{alg:cap}. The algorithm begins with an initialization of the hyperparameters - exploration-exploitation trade-off factor $\epsilon$, learning rate $\alpha$, discount factor $\gamma$, and discretization of the state space ($N$). For a desired $\delta_{\text{opt}}$, the budget is controlled by fixing the episodes, maximum allowable epochs ($n_{epoch}$), and two tolerance parameters, viz. tol$_{\delta}$ and tol$_{r}$. tol$_{\delta}$ is used to ensure convergence for each episode, and tol$_r$ for the acceptable error in the computed optima. The action taken by the agent is either rewarded or penalized ($r_t$) based on tol$_r$. A reward is assigned as $\frac{1}{\mid\delta_{\text{opt}} - \Delta \delta\mid}$ for $\Delta \delta < tol_r$, else a penalty is incurred as $-\mid\delta_{\text{opt}} - \Delta \delta\mid$. Here $\Delta \delta$ is the difference between the desired depth and predicted depth. This reward formulation ensures that $r_t$ is proportional to the accuracy of the optima.

\begin{algorithm}
\caption{Q-learning}\label{alg:cap}
\begin{algorithmic}[1]
\State  Initialize  \hspace{0.1in} $Q(s,a) = 0,\ \forall s \in \mathbb{S},\ a \in \mathbb{A}$ 
\State Initialize \hspace{0.1in} $\epsilon,\ \alpha,\ \gamma,\ $tol$_\delta$,\ $\delta_{\text{opt}},\ N,$ episodes, tol$_r$,\ $n_{\text{epochs}}$
\State Repeat for each episode
\Indent
\State Initialize $s_t$ arbitrarily
\State epochs = 0
\While {$\Delta \delta$ > tol$_\delta$ \& epochs $\leq$ $n_{\text{epochs}}$} : \Comment{$\Delta \delta = \mid \delta_{s_{t+1}}-\delta_{\text{opt}}\mid$}
\If{$\mathcal{U}_{[0,1]}<\epsilon$}
    \State Choose $a_t$ arbitrarily
\Else
    \State $a_t = \displaystyle\max_{a}Q(s_t,a)$ 
\EndIf
\State Take $a_t$, go to $s_{t+1}$
\State Evaluate $\delta(s_{t+1})$
\If{$\Delta \delta<$ tol$_r$}
    \State $r_t = \frac{1}{\mid\delta_{\text{opt}} - \Delta \delta\mid}$
\Else
    \State $r_t = -\mid\delta_{\text{opt}}-\Delta \delta\mid$
\EndIf
\State $ Q(s_t,a_t) \gets Q(s_t,a_t) + \alpha\left[r_t+\gamma \displaystyle\max_{a}Q(s_{t+1},a)-Q(s_t,a_t)\right] $
\State $s_t\gets s_{t+1} ;$
\State epochs $+=$ 1
\State until $s_t$ is terminal
\EndWhile
\EndIndent
\end{algorithmic}
\end{algorithm}

\subsection{Eagar-Tsai Environment} \label{Eagar-Tsai}

\begin{figure}
    \centering
    \includegraphics[width=0.9\textwidth]{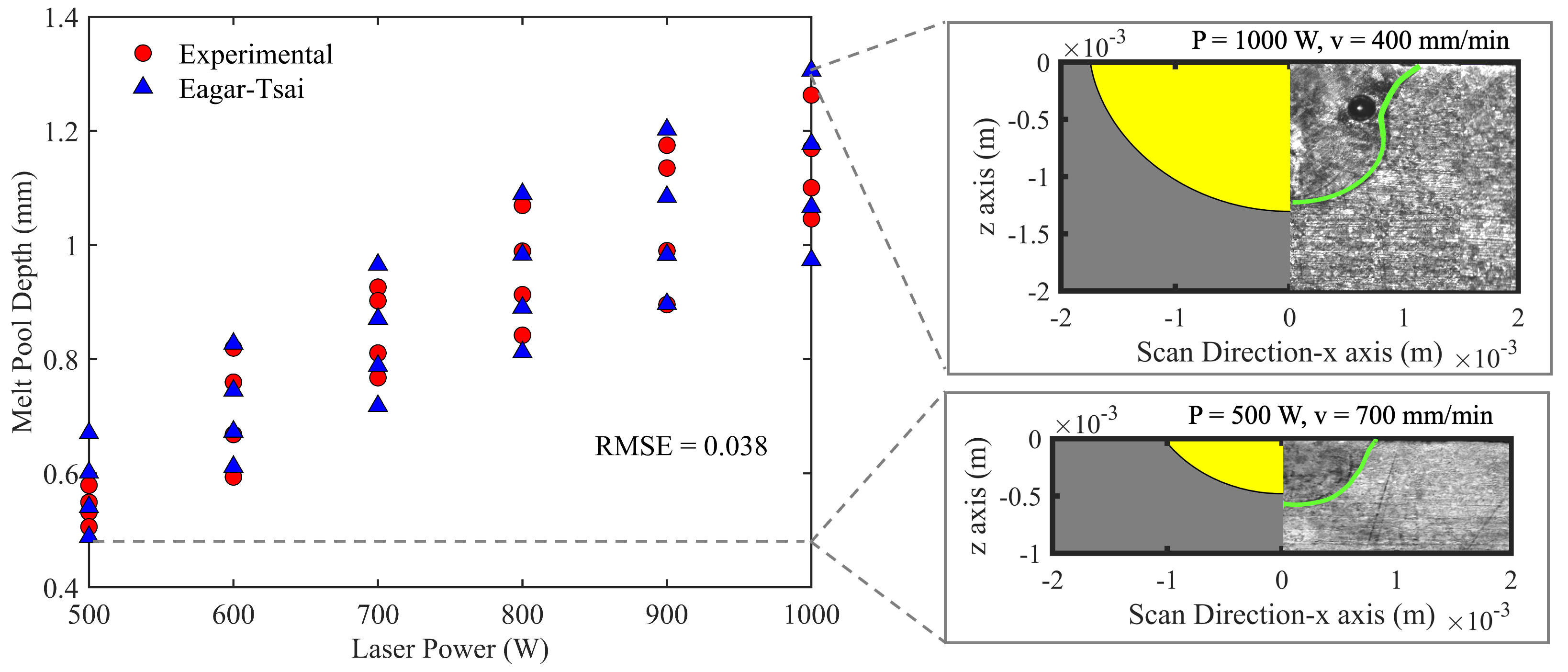}
    \caption{Validation of the Eagar-Tsai analytical function with experimental results. The figure on the left shows the two sources of $\delta$ compared to each other for every case of $P,v$ tested experimentally. The yellow portion corresponds to the liquid zone of the melt pool and the gray zone corresponds to the solid substrate. The calibrated Eagar-Tsai function shows a close agreement to the experimental results with a $RMSE = 0.038$. The two extended figures show the analytical melt pool juxtaposed with the experimental melt pool cross sections. The dashed yellow lines trace the melt pool boundary. The top figure corresponds to the largest $\delta$ obtained at $P = 1000 \ W, v = 400 \ mm/min$ corresponding to an actual $\delta$ of $1.262 \ mm$. The bottom figure corresponds to the shallowest melt pool for $P = 500 \ W, v = 700 \ mm/min$ corresponding to an actual depth of $0.506 \ mm$. }
    \label{fig:et_val}
\end{figure}

To provide physically-realistic estimates of $\delta$ over the state space i.e. the $P-v$ space of the L-DED system, the widely implemented thermal formulation developed by Eagar and Tsai~\cite{eagar1983temperature} is used. The function calculates the three-dimensional temperature distribution developed over a semi-infinite domain, produced by a Gaussian heat source. The temperature $T(x,y,z,t)$, at a particular location $(x,y,z)$ and time $t$ for a laser power $P$ and scan velocity $v$ in the $x$-direction is calculated as:
\begin{equation}
\begin{split}
   T(x,y,z,t) - T_0 =\frac{\alpha_L P}{\pi {\rho}_{p} c_p(4\pi a_p)^{1/2}} \int_{0}^{t} \frac{dt'(t-t')^{-1/2}}{2a_p(t-t')+{\sigma_L}^2} 
   e^{-\frac{(x-vt')^2+y^2}{4a_p(t-t')+2{\sigma_L}^2}-\frac{z^2}{4a_p(t-t')}}
\end{split}
\end{equation}
\begin{table}[h]
    \centering
    \caption{Thermal properties for SS316L and Eagar-Tsai function parameters}
    \begin{tabular}{c|c|c}
    \hline
        Parameter & Value & \\ 
    \hline
         Initial temperature of substrate, $T_0$ & 300 & $K$\\
         Liquidus temperature & 1700 & $K$\\
         Heat capacity, $c_p$ & 680 & $J/kg K$\\
         Density, $\rho_p$    & 7400 & $kg m^{-3}$\\
         Thermal diffusivity, $a_p$ & 7.1542 $\times 10^{-6}$ & $m^2 s^{-1}$\\
         Distribution parameter, $\sigma_L$ & 0.918 & - \\
         Absorptivity, $\alpha_L$ & 0.3 & - \\
    \hline
    \end{tabular}
    \label{tab:tab1}
\end{table}
Here, $T_0$ is the initial temperature of the substrate, $P$ is the laser power, $v$ is the scan velocity, $\alpha_L$ is the absorptivity of the laser beam, $\rho_p$ is the material density, $c_p$ is the specific heat, $a_p$ is the thermal diffusivity and $t^{\prime}$ is a dummy integration variable. $\sigma_L$ is the distribution parameter which is calibrated according to the experimental setup. 
The liquidus temperature acts as the melt pool boundary in the 3D temperature space and hence can be used to measure the steady-state melt pool depths. SS316L, a low-carbon steel alloy, is selected as the candidate material. The Eagar-Tsai formulation is calibrated using experimental melt pool depths measured for single-line, single-track deposits of SS316L powder (Osprey\textsuperscript{\textregistered} 316L provided by Sandvik) on a substrate of SS304. The experiments are carried out on a Meltio M450 L-DED system for laser power varying from 500 W to 1000 W, in steps of 100 W. For each case of $P$, $v$ is varied from 400 mm/min to 700 mm/min, increased in steps of 100 mm/min. The powder feed rate is maintained constant at 5L/min. Argon at a pressure of 4 bar is used as the carrier and shielding gas. The $\delta$s are measured by sectioning the deposits, characterizing the Bakelite-mounted specimens using standard metallographic preparation techniques for SS316L, and finally imaging via optical microscopy.  The results of the calibrated Eagar-Tsai function are shown in Figure. \ref{fig:et_val} where the simulated $\delta$s from the analytical function are plotted with the experimental $\delta$s for each case of $P,v$. Representative melt pools extracted from the analytical function are juxtaposed with their experimental counterpart in the two adjacent figures, where the top one is evaluated at the highest energy density ($P = 1000 \ W, v = 400 \ mm/min$) and the bottom one corresponds to the lowest energy density ($P = 500 \ W, v = 700 \ mm/min$). The thermal properties of SS316L and all other function parameters are included in Table. \ref{tab:tab1}. Evaluation of the function at a single ($P,v$) combination took $\sim 2$ seconds on an Intel\textsuperscript{\textregistered} Xeon\textsuperscript{\textregistered} Gold 6230. The analytical function is therefore a justified source of low-fidelity melt pool depths that can be sampled, as required, at significantly low cost of computation. This enables an efficient metric for developing a proof of concept for the present technique. The Eagar-Tsai simulations and the subsequent Q-learning algorithms are all developed on MATLAB\textsuperscript{\textregistered}.
\section{Results and Discussions}

\subsection{Q-learning performance}

\begin{figure}
    \centering
    \includegraphics[width=0.9\textwidth]{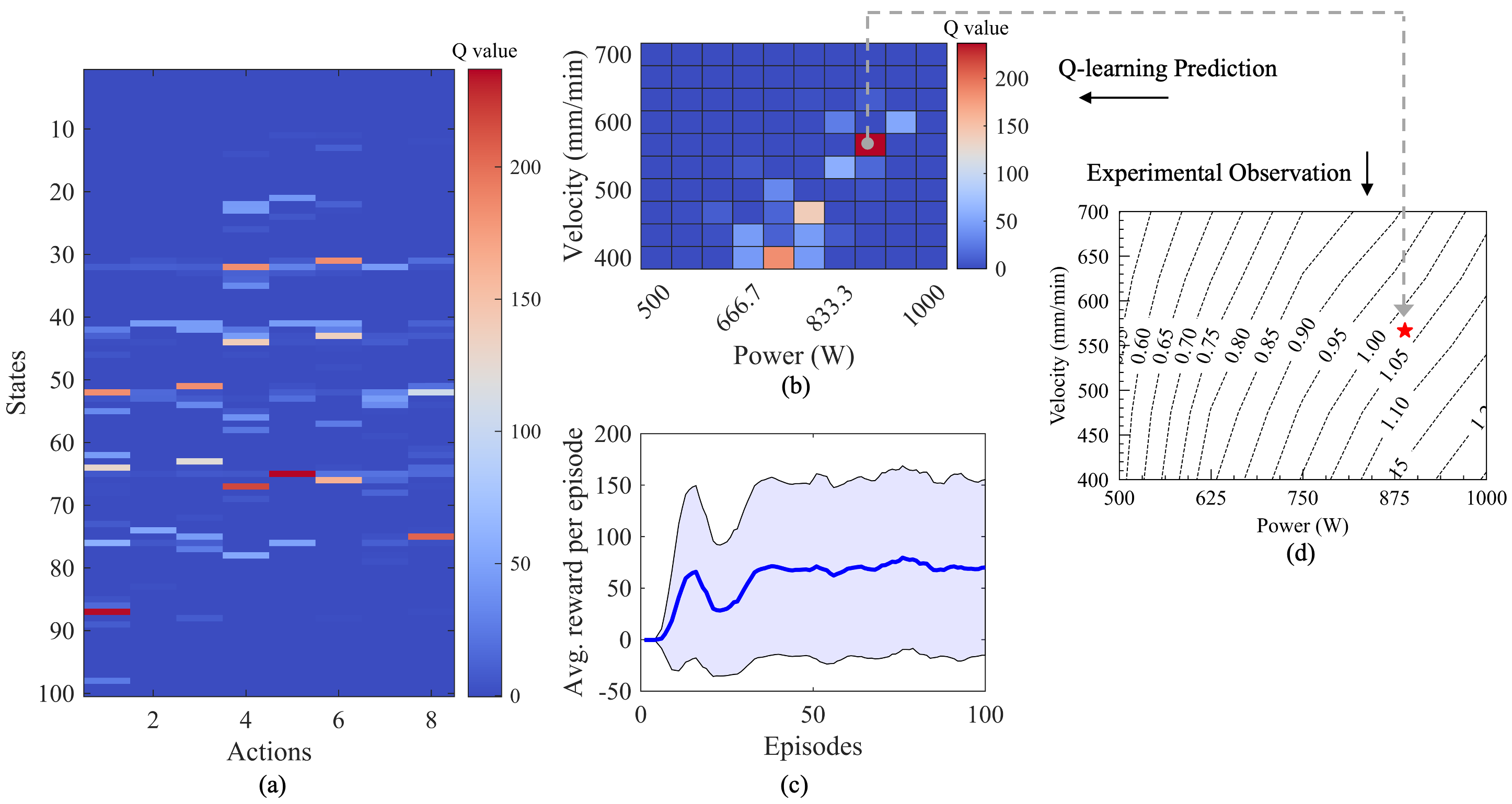}
    \caption{The performance of the Q-learning algorithm visualized through (a) Q-table. (b) state map. (c) convergence curve, and (d) experimental validation. The state map reveals the optimum $P-v$ combination as $P = 888.9 \ W$ and $v = 566.7 \ mm/min$ leading to $\delta = 1.0045 \ mm$. This result also shows a correlation to the experimentally obtained $P-v$ map in (d).}
    \label{fig:qperf}
\end{figure}

The algorithm is simulated for a space discretized into 100 states with $N = 10$, and with the parameters $\alpha = 0.25$, $\epsilon = 0.25$, and $\gamma = 0.25$. The results at the end of 100 episodes for optimizing the melt pool depth (${\delta}_{opt} = 1 \ mm$) is shown in Figure. \ref{fig:qperf}. The Q-table (Figure. \ref{fig:qperf}(a)) shows the Q-values accumulated at each state-action pair. The state-action pairs with the high Q-values correspond to $P-v$ combinations in the process parameter space that yield a $\delta$ closer to ${\delta}_{opt}$. A better visualization of the optimal $P-v$ combinations is revealed by a mapping from the Q-table to the $P-v$ space as shown in Figure. \ref{fig:qperf}(b). A power $P = 888.9 \ W$, and velocity $v = 566.7 \ mm/min$ is observed to possess the maximum Q-value resulting in a $\delta$ of $1.0045 \ mm$. In addition to the optimal solution, the figure also reveals multiple $P-v$ combinations that have relatively high Q-values. Particularly, $P = 722.2 \ W,\ v = 400 \ mm/min$, and $P = 777.78 \ W,\ v = 466.67 \ mm/min$, are observed to be the next best solutions with a $\delta$ of $0.994 \ mm$ and $0.992 \ mm$ respectively. The average rewards per episode with an uncertainty band is shown in Figure. \ref{fig:qperf}(c). During an episode, the algorithm explores high reward and high penalty regimes incurring a large variance in rewards through the entire learning process. This is mainly driven by the parameter $\epsilon$ which ensures an exploration of the state space even after the knowledge of a relatively well performing optimal solution.
A process map, as shown in Figure. \ref{fig:qperf}(d), is derived from the experimental results discussed in section~\ref{Eagar-Tsai}. The map plots contours of constant $\delta$ over the predefined process parameter space and, thereby, ensures a calibration of the prediction from the Q-learning process. Figure. \ref{fig:qperf}(d) shows a superimposition of the optimal solution from Q-learning onto this process map. A physical correlation to the experimental $1 \ mm$ contour can be discerned from Figure. \ref{fig:qperf}(b) where the relatively higher Q-values trace a set of $P-v$ combinations that also correspond to a similar contour. The algorithm, therefore, not only computes a point estimate but a set of plausible solutions.

\subsection{Effect of hyperparameters}

There are five hyperparameters that can influence the behavior of the Q-learning algorithm. They are, (i) domain discretization ($N$), (ii) exploration-exploitation tradeoff parameter ($\epsilon$), (iii) discount factor ($\gamma$), (iv) learning rate ($\alpha$), and (v) number of episodes. This section presents the impacts of each of these parameters. For all the cases, the default parameters are maintained at $N = 10$, $\epsilon, \gamma, \alpha = 0.25$, and episodes = 100 unless varied.

\subsubsection{Effect of discretization ($N$)}

\begin{figure}
    \centering
    \includegraphics[width=0.9\textwidth]{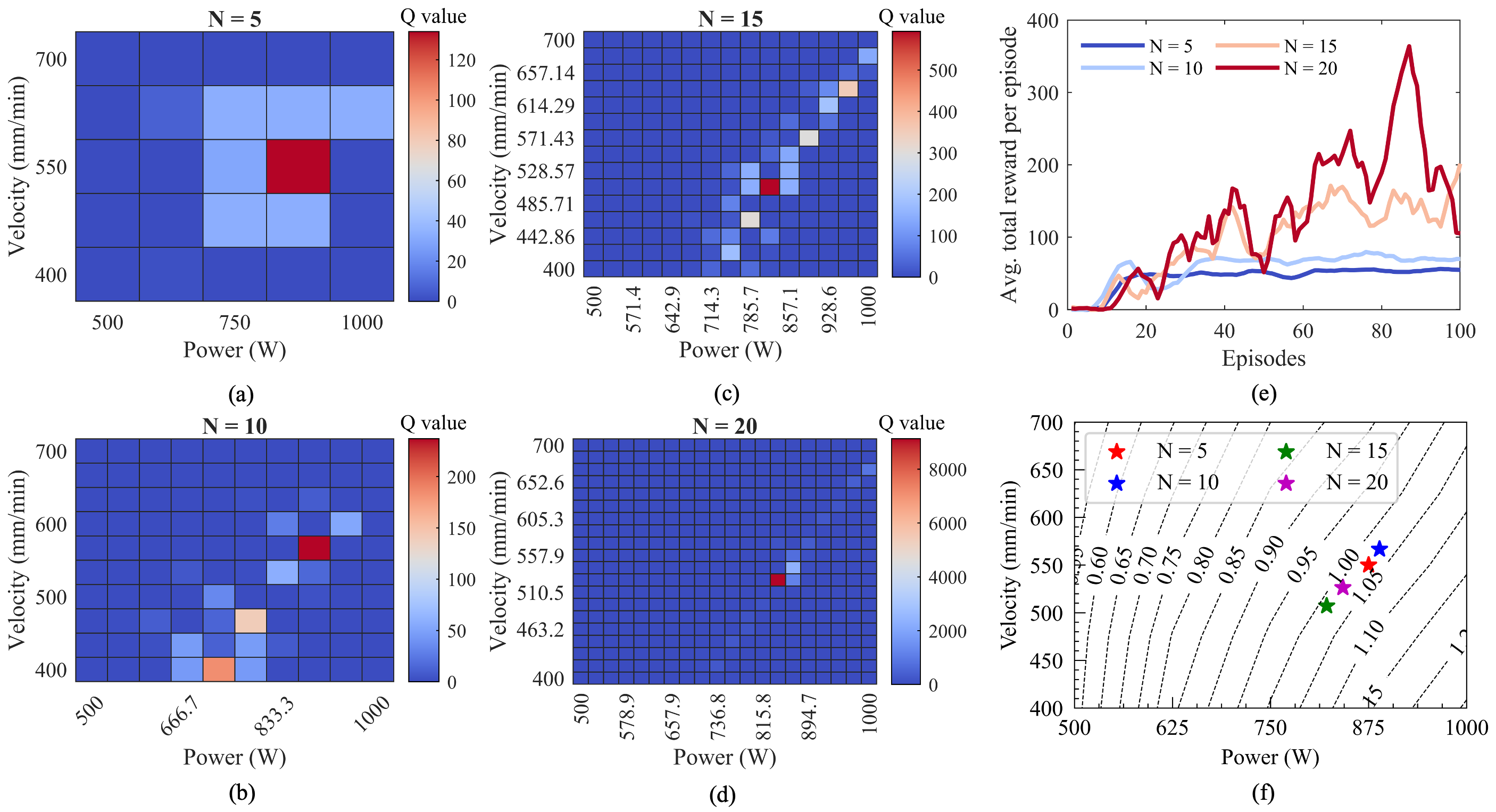}
    \caption{The effect of discretization on Q-learning shown through $P-v$ maps for (a) $N = 5$, (b) $N = 10$, (c) $N = 10$, and (d) $N = 20$. (e) A comparison of the convergence curves. (f) The optimal points overlaid on the experimental $P-v$ map.}
    \label{fig:hyp_n}
\end{figure}

The discretization of the $P-v$ space, defined with the parameter $N$, is determined by the desired resolution and accuracy of the solution. Four cases with N varying from 5 to 20 have been considered to study the impact. Figure. \ref{fig:hyp_n} summarizes the results through corresponding $P-v$ maps ((a)-(d)), convergence of rewards (e), and the superimposition of the optimal points on the experimentally derived process map (f). The optimal solutions obtained for $N = 5, 10, 15$, and $20$ are 1.008 $mm$, 1.0045 $mm$, 0.9982 $mm$, and 0.9999 $mm$, respectively. The Q-values and their variabilities from the $P-v$ maps are observed to be directly proportional to $N$. $N=20$ displays the highest Q-values, and $N = 5$ displays the lowest. Moreover, for $N = 10, 15,$ and $20$, a distinct contour of relatively higher Q-values is observed. These correspond to a set of $P-v$ combinations that yield a $\delta$ closer to ${\delta}_{opt}$. Furthermore, the convergence plot (Figure. \ref{fig:hyp_n}(e)) indicates an inverse proportionality to $N$. The final rewards show that $N = 15, 20$ incur more rewards than $N = 5, 10$. This is because a coarser space forces the Q-learning algorithm to exhaust the possible $P-v$ combinations it can take to maximize rewards. The final optimal solutions for each case are overlaid on the experimental process map in Figure. \ref{fig:hyp_n}(f). All the points lie close to the $1 \ mm$ contour indicating that the Q-learning algorithm is robust to the discretization of the $P-v$ space.

\subsubsection{Effect of the exploration-exploitation trade-off parameter $\epsilon$}

\begin{figure}
    \centering
    \includegraphics[width=0.9\textwidth]{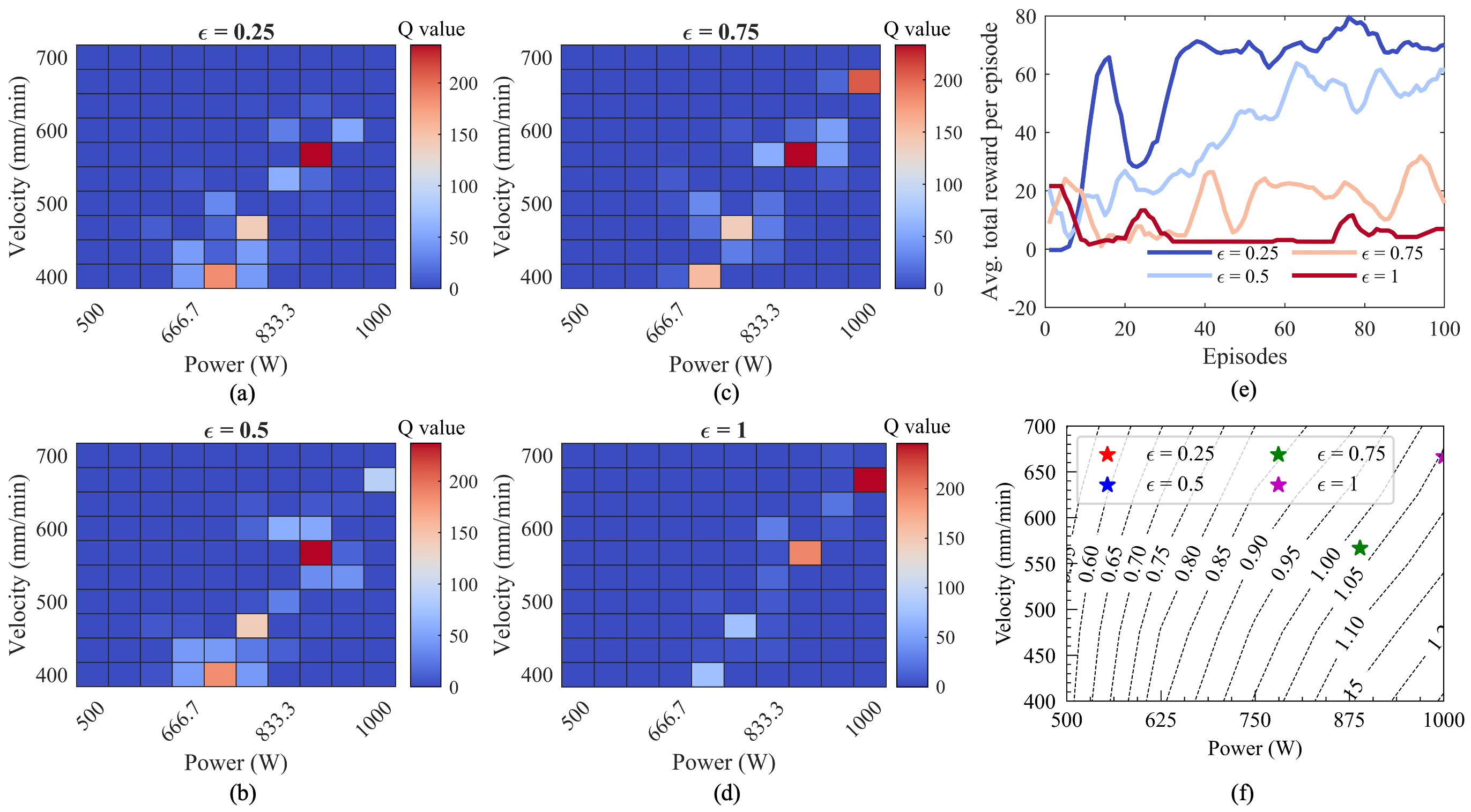}
    \caption{The effect of the exploration-exploitation trade-off parameter $\epsilon$ on Q-learning shown through $P-v$ maps for (a) $\epsilon = 0.25$, (b) $\epsilon = 0.5$, (c) $\epsilon = 0.75$, and (d) $\epsilon = 1$. (e) A comparison of the convergence curves. (f) The optimal points overlaid on the experimental $P-v$ map.}
    \label{fig:hyp_eps}
\end{figure}

The trade-off between the choice of the agent to \textit{explore} or \textit{exploit} is dictated by the hyperparameter $\epsilon$. In Algorithm \ref{alg:cap}, $\epsilon$ is implemented in lines 7-11. A high value of $\epsilon$ drives exploration where the algorithm can take a random action to probe the environment. A low value of $\epsilon$ favors exploitation which makes the Q-learning algorithm take an action based on the existing Q-values. Four different cases with $\epsilon = 0.25, 0.5, 0.75,$ and $1$ are studied. The corresponding results are shown in Figure. \ref{fig:hyp_eps}. Figures. \ref{fig:hyp_eps}(a)-(d) show the P-v maps for all four cases, Figure. \ref{fig:hyp_eps}(e) shows the convergence rates and Figure. \ref{fig:hyp_eps}(f) traces the results to the experimental $P-v$ map. Among all the $P-v$ maps, more points with high Q-values are observed around $\delta = 1 \ mm$ contour for $\epsilon = 0.25, 0.5,$ and $0.75$. This is due to the exploitation tendency for all values of $\epsilon$ except $\epsilon = 1$ which is pure exploration. In Figure. \ref{fig:hyp_eps}(e), for lower $\epsilon$, the amount of rewards accrued grows at a faster rate due to its probability to take actions that maximize rewards. For the extreme case with $\epsilon = 1$, a flat line is observed barring a few chance episodes where the algorithm encounters a good solution. There is no mechanism for this case to gravitate further towards a better optimum. A comparison with the experimental $P-v$ map shows the optimal solution from $\epsilon = 1$ to be farther from the actual result as compared to the other three coincidental optima. However, the optimal $P-v$ obtained from $\epsilon = 1$ corresponds to an analytical prediction of $\delta = 1.0028$ as compared to $\delta = 1.0045 \ mm$ for $\epsilon = 0.25, 0.5, 0.75$. This discrepancy can be attributed to the shortcomings of the analytical Eagar-Tsai environment.

\subsubsection{Effect of the discount factor $\gamma$}
\begin{figure}
    \centering
    \includegraphics[width=0.9\textwidth]{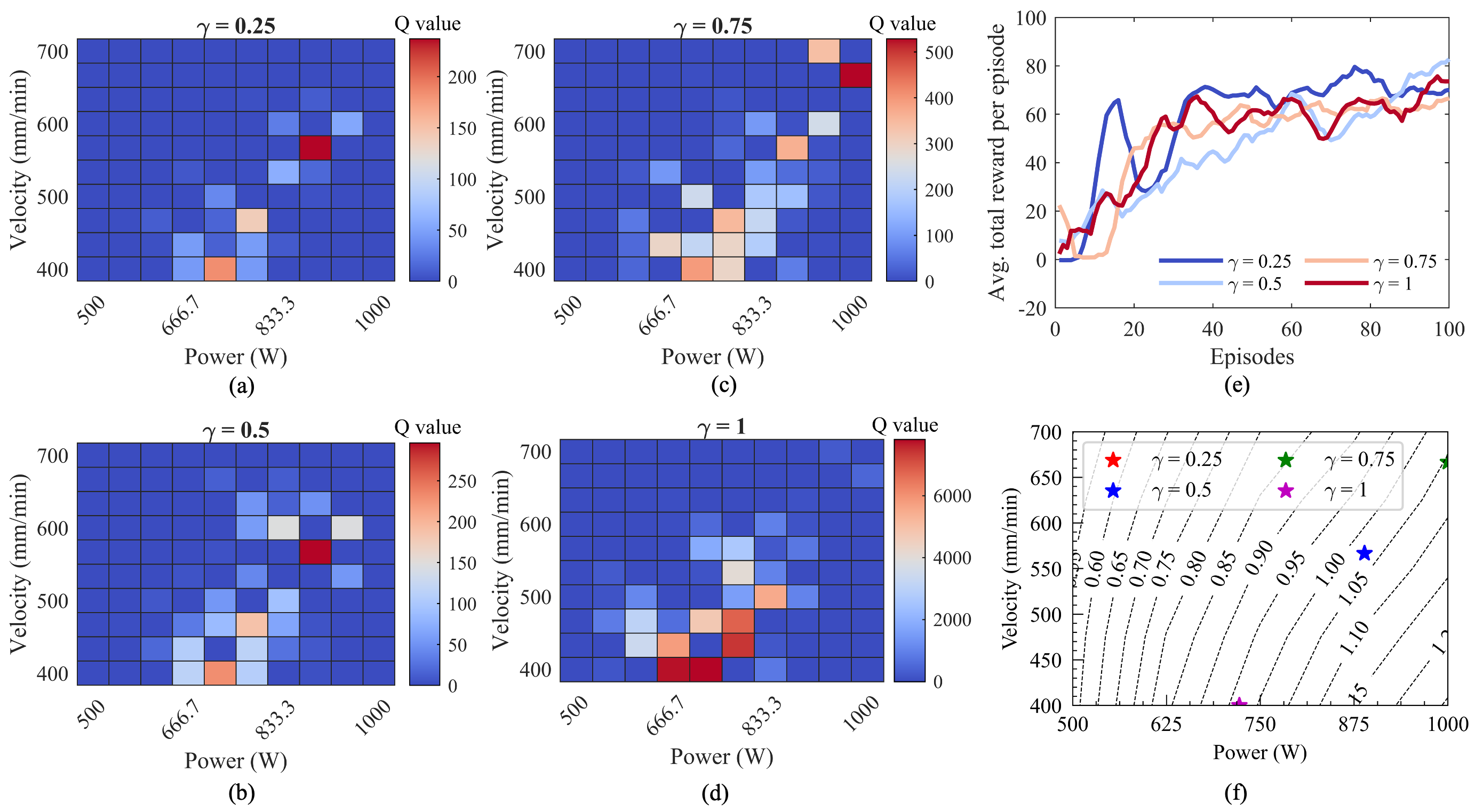}
    \caption{The effect of the discount factor $\gamma$ on Q-learning shown through $P-v$ maps for (a) $\gamma = 0.25$, (b) $\gamma = 0.5$, (c) $\gamma = 0.75$, and (d) $\gamma = 1$. (e) A comparison of the convergence curves. (f) The optimal points overlaid on the experimental $P-v$ map.}
    \label{fig:hyp_gamma}
\end{figure}

The discount rate, $\gamma$, is used to control the trade-off between long-term and immediate rewards. A higher $\gamma$ assigns more importance to future rewards while a lower $\gamma$ makes the Q-learning algorithm greedy. Mathematically, $\gamma$ has a proportional impact on the Q-value, as captured by Eq.\ref{eq}. Figure. \ref{fig:hyp_gamma} shows the results for four cases of $\gamma$. Higher values of $\gamma$ tend to yield more points with high Q-values in the $P-v$ maps. This is because the algorithm assigns importance to more lucrative prospects. The convergence curves in Figure. \ref{fig:hyp_gamma}(e) do not show any conclusive correlation between the average total rewards and $\gamma$ values. The significantly high Q-values for $\gamma = 1$ in contrast to the corresponding average rewards are therefore indicative of the algorithm being directed to the same set of $P-v$ combinations that maximize the rewards. This disparity is due to the interplay of two phenomena, (i) the proportionality between $\gamma$ and Q-value (Eq.\ref{eq}) and (ii) Q-value's influence on the choice of action (Algorithm \ref{alg:cap} line 10). The experimental mapping reveals that $\gamma = 0.25$ and $0.5$ have the best (coincidental) optima.

\subsubsection{Effect of the learning rate $\alpha$}
\begin{figure}
    \centering
    \includegraphics[width=0.9\textwidth]{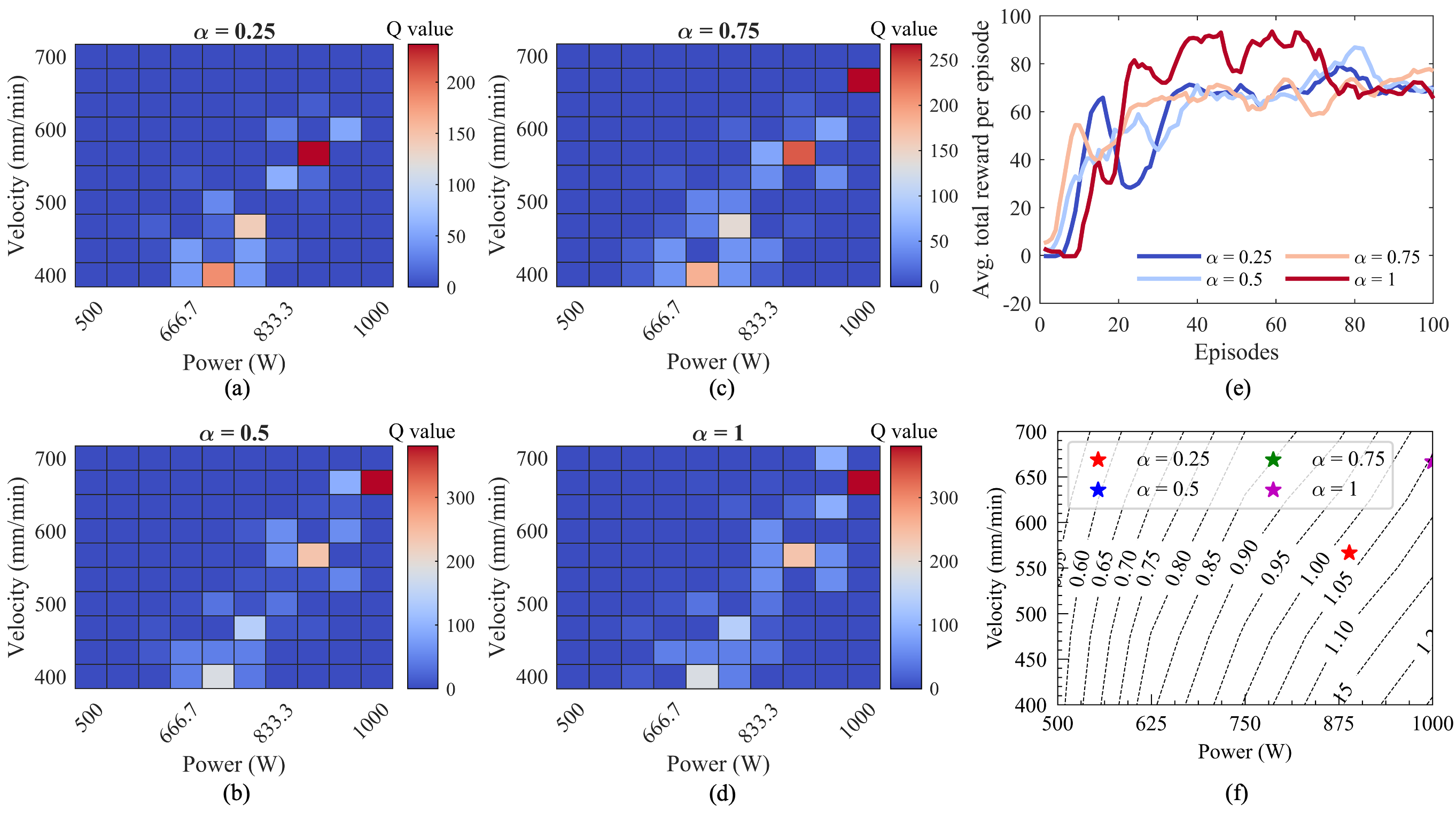}
    \caption{The effect of the learning rate $\alpha$ on Q-learning shown through $P-v$ maps for (a) $\alpha = 0.25$, (b) $\alpha = 0.5$, (c) $\alpha = 0.75$, and (d) $\alpha = 1$. (e) A comparison of the convergence curves. (f) The optimal points overlaid on the experimental $P-v$ map.}
    \label{fig:hyp_alpha}
\end{figure}

The learning rate, $\alpha$, controls the rate at which the Q-value is updated during every epoch. $\alpha$ can take values between `0' and `1'; `0' makes the learning redundant as the agent relies only on prior knowledge, while a value of `1' makes the agent consider only the most recent information, ignoring prior knowledge (Eq.\ref{eq}). The four different $P-v$ maps obtained through the variation of $\alpha$ are shown in Figure. \ref{fig:hyp_alpha}(a)-(d). As compared to the previous three hyperparameters, the impact of $\alpha$ on the learning tendency of the algorithm is not consequential. No significant difference is observed between the total rewards and the convergence curves (Figure. \ref{fig:hyp_alpha} (e)). The same optimal points are obtained for $\alpha = 0.5, 0.75,$ and 1. This optimum is farther from ${\delta}_{opt}$ as compared to the optima for $\alpha = 0.25$ as observed from the experimental P-v map in (Figure. \ref{fig:hyp_alpha}(f)). Therefore the use of $\alpha = 0.25$ is justified for a successful application of the algorithm.

\subsubsection{Effect of number of episodes}
\begin{figure}
    \centering
    \includegraphics[width=0.9\textwidth]{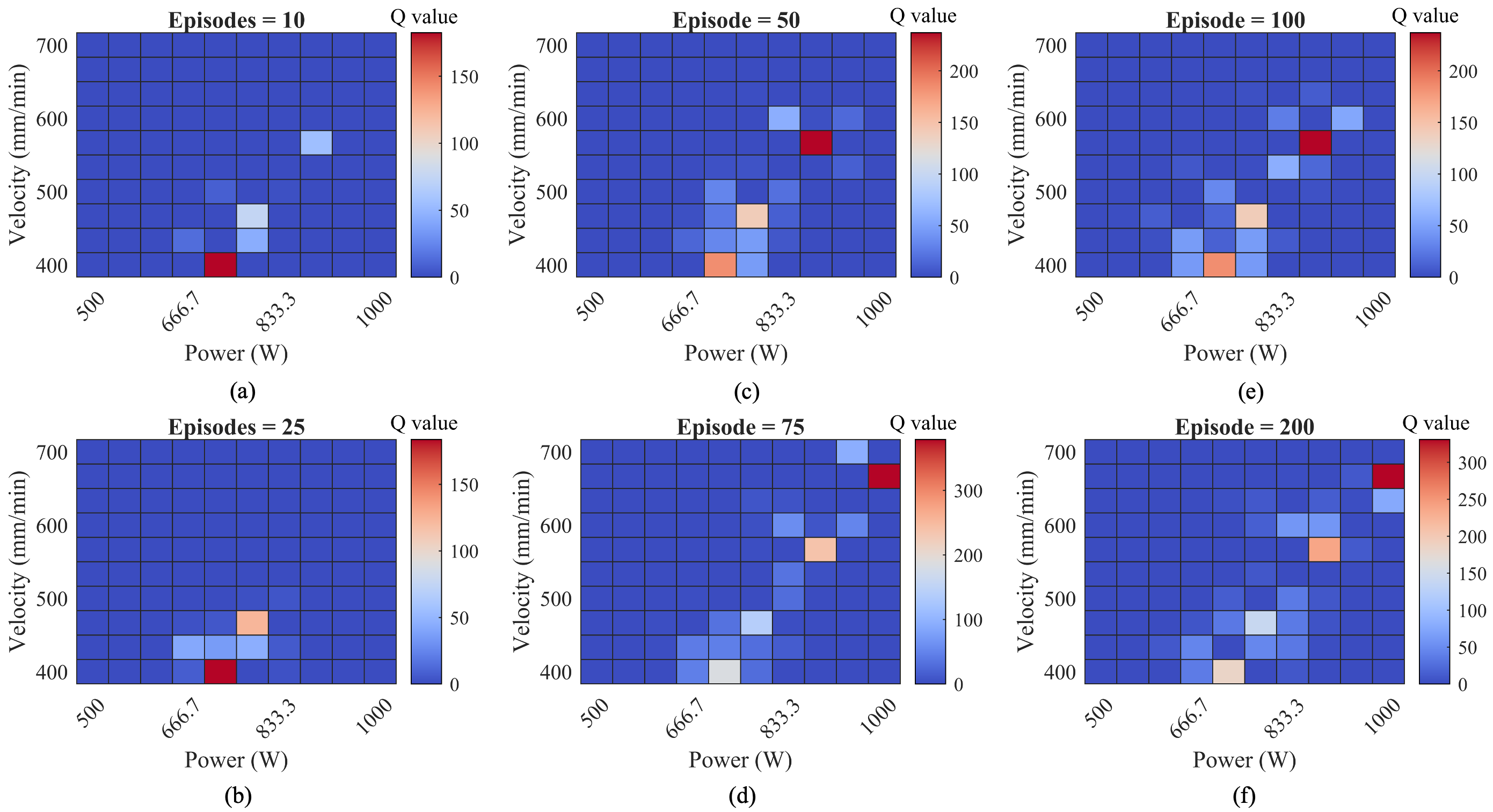}
    \caption{The effect of number of episodes on Q-learning shown through $P-v$ maps for (a) 10, (b) 25, (c) 50, (d) 75, (e) 100, and (f) 200 episodes.}
    \label{fig:hyp_epsd}
\end{figure}

In a realistic scenario where the interaction is directly with any AM hardware as opposed to a simulated function such as Eagar-Tsai, it may still be practically impossible to conduct an optimization with 100 episodes as demonstrated above. However, the reasoning behind developing this concept for AM is to highlight its capability in estimating \textit{some} intelligently informed solution to the problem at any stage. To further understand this capability, the quality of the Q-learning solution is now studied for varying number of episodes ranging from 10 to 200. The evolving nature of the $P-v$ maps for all the episodes is shown in Figure. \ref{fig:hyp_epsd}. Even for 10 and 25 training episodes, the predicted optimal solution leads to $\delta = 0.994 \ mm$. This exemplifies the mechanisms of Q-learning to provide a solution with scarce data. Understandably, the optimum prediction improves for 50 episodes where $\delta = 1.0045 \ mm$, and for 75, 100, and 200 episodes, $\delta = 1.0028 \ mm$. With these observations (and from preceding sections), Q-learning presents a strong candidate for further process parameter optimization studies for various scenarios in AM.

\section{Conclusion and future work}

The study tackles a relatively unexplored process parameter optimization problem with an on-the-fly model-free Q-learning based RL. The optimization is targeted towards finding a set of $P$ and $v$ combinations such that $\delta$ is maintained at 1 $mm$. To set up the RL framework for this AM problem, the interpretation of \textit{agent}, \textit{environment}, \textit{state space}, and \textit{action space} is presented. Accordingly, the laser is compared to an agent whereas the Eagar-Tsai function is used as the environment. The results from the optimization reveal an optimal estimate of the $P,v$ combination that can maintain $\delta = 1.0045 \ mm$. A successful comparison of this optimal estimate with an experimentally derived process map is also discussed. To further dissect the working mechanisms of Q-learning, a detailed study of five hyperparameters, viz. (i) discretization (N), (ii) exploration-exploitation parameter ($\epsilon$), (iii) discount factor ($\gamma$), (iv) learning rate ($\alpha$), and (v) number of episodes, is conducted. The study, therefore, presents a viable alternative to process parameter optimization in challenging scenarios where there is either no system information or low data available.

Owing to some recent (and ongoing) developments in RL, there are multiple ways in which the current study can be further developed. To begin with, a discrete parameter space provides access to only certain combinations from the design space. This may not be favorable in situations where the best $P-v$ combination may have been lost due to discretization. To tackle this problem, there are alternatives in RL such as proximal policy optimization \cite{schulman2017proximal} that uses a continuous parameter space. The discretized domain, however, is a better alternative to quickly sample the space for a good initial condition with limited information. In addition to discretization, the optimization can begin with a better (or informed) initial formulation for the Q-table. As a hypothetical scenario, this study began the optimization with a null Q-table. In practice, the manufacturer can provide certain recommendations for favorable parameter combinations. These recommendations can be reflected in the Q-table by assigning higher initial Q-values to further accelerate the search for a better optimum.

There can also be multiple ways in which the formulation of the \textit{environment} and \textit{action space} can be tackled. Although Eagar-Tsai has been used in this problem, higher-fidelity alternatives such as finite element and computational models can be employed to provide better estimates. Online interaction with an L-DED system, the ulterior motive of developing this research, is also feasible albeit with some concurrent material characterization. The action space provides some interesting research and development directions. For this problem, since only two parameters ($P, v$) were considered, the action space was limited to eight distinct actions. With further increase in parameters, this action space will increase exponentially leading to an intractable problem. This is also inevitable due to the multitude of process parameters that are commonly associated with AM. In such conditions, multi-agent reinforcement learning (MARL) \cite{zhang2021multi} can prove to be a viable alternative. Broadly, MARL treats every parameter as a separate agent and provides an individual Q-table eliminating the need for manual selection of the action space. The RL literature, in summary, has multiple potent alternatives to further enrich the process parameter optimization research.

\section*{Credit Authorship}
Conceptualization, N.M. and S.D.; methodology, N.M. and S.D.; software, N.M. and S.D.; validation, N.M. and S.D.; formal analysis, N.M. and S.D.; investigation, N.M. and S.D.; resources, A.B.; data curation, N.M. and S.D.; writing---original draft preparation, N.M. and S.D.; writing---review and editing, N.M., S.D. and A.B.; visualization, N.M. and S.D.; supervision, A.B.; project administration, A.B.; funding acquisition, A.B. All authors have read and agreed to the published version of the manuscript.

\section*{Funding Information}
The research is funded in part by the NASA University Leadership Initiative program through grant number 80NSSC21M0068 and in part by the U.S. Army Engineer Research and Development Center through Contract Number W912HZ21C0001. Any opinions, findings, and conclusions in this paper are those of the authors and do not necessarily reflect the views of the supporting institution.

\section*{Data Availability Statement}
The data is available from the communicating author on reasonable request.

\section*{Conflicts of Interest}
The authors declare no conflict of interest.

 \bibliographystyle{unsrt} 
 \bibliography{V1}

\begin{thebibliography}{10}

\bibitem{debroy2018additive}
Tarasankar DebRoy, HL~Wei, JS~Zuback, Tuhin Mukherjee, JW~Elmer, JO~Milewski,
  Allison~Michelle Beese, A~De Wilson-Heid, Amitava De, and W~Zhang.
\newblock Additive manufacturing of metallic components--process, structure and
  properties.
\newblock {\em Progress in Materials Science}, 92:112--224, 2018.

\bibitem{ferster2018effects}
Katharine~K Ferster, Kathryn~L Kirsch, and Karen~A Thole.
\newblock Effects of geometry, spacing, and number of pin fins in additively
  manufactured microchannel pin fin arrays.
\newblock {\em Journal of Turbomachinery}, 140(1), 2018.

\bibitem{farshidianfar2016effect}
Mohammad~H Farshidianfar, Amir Khajepour, and Adrian~P Gerlich.
\newblock Effect of real-time cooling rate on microstructure in laser additive
  manufacturing.
\newblock {\em Journal of Materials Processing Technology}, 231:468--478, 2016.

\bibitem{vasinonta2000process}
Aditad Vasinonta, Jack Beuth, and Michelle Griffith.
\newblock Process maps for controlling residual stress and melt pool size in
  laser-based sff processes 200.
\newblock In {\em 2000 International Solid Freeform Fabrication Symposium},
  2000.

\bibitem{basak2016additive}
Amrita Basak, Ranadip Acharya, and Suman Das.
\newblock Additive manufacturing of single-crystal superalloy cmsx-4 through
  scanning laser epitaxy: computational modeling, experimental process
  development, and process parameter optimization.
\newblock {\em Metallurgical and Materials Transactions A}, 47(8):3845--3859,
  2016.

\bibitem{berumen2010quality}
Sebastian Berumen, Florian Bechmann, Stefan Lindner, Jean-Pierre Kruth, and Tom
  Craeghs.
\newblock Quality control of laser-and powder bed-based {Additive Manufacturing
  (AM)} technologies.
\newblock {\em Physics procedia}, 5:617--622, 2010.

\bibitem{raghavan2013heat}
A~Raghavan, HL~Wei, TA~Palmer, and Tarasankar Debroy.
\newblock Heat transfer and fluid flow in additive manufacturing.
\newblock {\em Journal of Laser Applications}, 25(5):052006, 2013.

\bibitem{everton2016review}
Sarah~K Everton, Matthias Hirsch, Petros Stravroulakis, Richard~K Leach, and
  Adam~T Clare.
\newblock Review of in-situ process monitoring and in-situ metrology for metal
  additive manufacturing.
\newblock {\em Materials \& Design}, 95:431--445, 2016.

\bibitem{liao2022simulation}
Shuheng Liao, Samantha Webster, Dean Huang, Raymonde Council, Kornel Ehmann,
  and Jian Cao.
\newblock Simulation-guided variable laser power design for melt pool depth
  control in directed energy deposition.
\newblock {\em Additive Manufacturing}, page 102912, 2022.

\bibitem{gockel2013understanding}
Joy Gockel and Jack Beuth.
\newblock Understanding {Ti-6Al-4V} microstructure control in additive
  manufacturing via process maps.
\newblock In {\em 2013 International Solid Freeform Fabrication Symposium}.
  University of Texas at Austin, 2013.

\bibitem{bhardwaj2019direct}
Tarun Bhardwaj, Mukul Shukla, CP~Paul, and KS~Bindra.
\newblock Direct energy deposition-laser additive manufacturing of
  titanium-molybdenum alloy: Parametric studies, microstructure and mechanical
  properties.
\newblock {\em Journal of Alloys and Compounds}, 787:1238--1248, 2019.

\bibitem{aboutaleb2017accelerated}
Amir~M Aboutaleb, Linkan Bian, Alaa Elwany, Nima Shamsaei, Scott~M Thompson,
  and Gustavo Tapia.
\newblock Accelerated process optimization for laser-based additive
  manufacturing by leveraging similar prior studies.
\newblock {\em IISE Transactions}, 49(1):31--44, 2017.

\bibitem{velazquez2021prediction}
Daniel Ren{\'e}~Tas{\'e} Vel{\'a}zquez, Andr{\'e}~Lu{\'\i}s Helleno,
  Hip{\'o}lito~Carvajal Fals, and Raphael~Galdino dos Santos.
\newblock Prediction of geometrical characteristics and process parameter
  optimization of laser deposition {AISI} 316 steel using fuzzy inference.
\newblock {\em The International Journal of Advanced Manufacturing Technology},
  115(5):1547--1564, 2021.

\bibitem{lu2010prediction}
ZL~Lu, DC~Li, BH~Lu, AF~Zhang, GX~Zhu, and G~Pi.
\newblock The prediction of the building precision in the laser engineered net
  shaping process using advanced networks.
\newblock {\em Optics and Lasers in Engineering}, 48(5):519--525, 2010.

\bibitem{akbari2022meltpoolnet}
Parand Akbari, Francis Ogoke, Ning-Yu Kao, Kazem Meidani, Chun-Yu Yeh, William
  Lee, and Amir~Barati Farimani.
\newblock Meltpoolnet: Melt pool characteristic prediction in metal additive
  manufacturing using machine learning.
\newblock {\em Additive Manufacturing}, 55:102817, 2022.

\bibitem{mondal2020investigation}
Sudeepta Mondal, Daniel Gwynn, Asok Ray, and Amrita Basak.
\newblock Investigation of melt pool geometry control in additive manufacturing
  using hybrid modeling.
\newblock {\em Metals}, 10(5):683, 2020.

\bibitem{menon2022multi}
Nandana Menon, Sudeepta Mondal, and Amrita Basak.
\newblock Multi-fidelity surrogate-based process mapping with uncertainty
  quantification in laser directed energy deposition.
\newblock {\em Materials}, 15(8):2902, 2022.

\bibitem{wang2019data}
Zhuo Wang, Pengwei Liu, Yaohong Xiao, Xiangyang Cui, Zhen Hu, and Lei Chen.
\newblock A data-driven approach for process optimization of metallic additive
  manufacturing under uncertainty.
\newblock {\em Journal of Manufacturing Science and Engineering}, 141(8), 2019.

\bibitem{ye2018novel}
Zhipeng Ye, Jiawei Zhu, Qian Li, Bing Mo, Baimao Lei, Yaqiu Li, Chunhui Wang,
  and Chuangmian Huang.
\newblock A novel method of reliability-centered process optimization for
  additive manufacturing.
\newblock {\em Microelectronics Reliability}, 88:1151--1156, 2018.

\bibitem{kaelbling1996reinforcement}
Leslie~Pack Kaelbling, Michael~L Littman, and Andrew~W Moore.
\newblock Reinforcement learning: A survey.
\newblock {\em Journal of artificial intelligence research}, 4:237--285, 1996.

\bibitem{kober2013reinforcement}
Jens Kober, J~Andrew Bagnell, and Jan Peters.
\newblock Reinforcement learning in robotics: A survey.
\newblock {\em The International Journal of Robotics Research},
  32(11):1238--1274, 2013.

\bibitem{kaiser2019model}
Lukasz Kaiser, Mohammad Babaeizadeh, Piotr Milos, Blazej Osinski, Roy~H
  Campbell, Konrad Czechowski, Dumitru Erhan, Chelsea Finn, Piotr Kozakowski,
  Sergey Levine, et~al.
\newblock Model-based reinforcement learning for atari.
\newblock {\em arXiv preprint arXiv:1903.00374}, 2019.

\bibitem{silver2018general}
David Silver, Thomas Hubert, Julian Schrittwieser, Ioannis Antonoglou, Matthew
  Lai, Arthur Guez, Marc Lanctot, Laurent Sifre, Dharshan Kumaran, Thore
  Graepel, et~al.
\newblock A general reinforcement learning algorithm that masters chess, shogi,
  and go through self-play.
\newblock {\em Science}, 362(6419):1140--1144, 2018.

\bibitem{fawzi2022discovering}
Alhussein Fawzi, Matej Balog, Aja Huang, Thomas Hubert, Bernardino
  Romera-Paredes, Mohammadamin Barekatain, Alexander Novikov, Francisco~J
  R~Ruiz, Julian Schrittwieser, Grzegorz Swirszcz, et~al.
\newblock Discovering faster matrix multiplication algorithms with
  reinforcement learning.
\newblock {\em Nature}, 610(7930):47--53, 2022.

\bibitem{mozaffar2020toolpath}
Mojtaba Mozaffar, Ablodghani Ebrahimi, and Jian Cao.
\newblock Toolpath design for additive manufacturing using deep reinforcement
  learning.
\newblock {\em arXiv preprint arXiv:2009.14365}, 2020.

\bibitem{ogoke2021thermal}
Francis Ogoke and Amir~Barati Farimani.
\newblock Thermal control of laser powder bed fusion using deep reinforcement
  learning.
\newblock {\em Additive Manufacturing}, 46:102033, 2021.

\bibitem{zimmerling2022optimisation}
Clemens Zimmerling, Christian Poppe, Oliver Stein, and Luise K{\"a}rger.
\newblock Optimisation of manufacturing process parameters for variable
  component geometries using reinforcement learning.
\newblock {\em Materials \& Design}, 214:110423, 2022.

\bibitem{watkins1992q}
Christopher~JCH Watkins and Peter Dayan.
\newblock Q-learning.
\newblock {\em Machine learning}, 8(3):279--292, 1992.

\bibitem{sutton2018reinforcement}
Richard~S Sutton and Andrew~G Barto.
\newblock {\em Reinforcement learning: An introduction}.
\newblock MIT press, 2018.

\bibitem{eagar1983temperature}
TW~Eagar, NS~Tsai, et~al.
\newblock Temperature fields produced by traveling distributed heat sources.
\newblock {\em Welding journal}, 62(12):346--355, 1983.

\bibitem{schulman2017proximal}
John Schulman, Filip Wolski, Prafulla Dhariwal, Alec Radford, and Oleg Klimov.
\newblock Proximal policy optimization algorithms.
\newblock {\em arXiv preprint arXiv:1707.06347}, 2017.

\bibitem{zhang2021multi}
Kaiqing Zhang, Zhuoran Yang, and Tamer Ba{\c{s}}ar.
\newblock Multi-agent reinforcement learning: A selective overview of theories
  and algorithms.
\newblock {\em Handbook of Reinforcement Learning and Control}, pages 321--384,
  2021.

\end{thebibliography}





\end{document}